\documentclass[aos,MSNbibl,nameyear,dvips]{arximspdf}

\doi{10.1214/14-AOS1217} 
\volume{42}
\issue{4}
\pubyear{2014}
\firstpage{1689}
\lastpage{1691}

\makeatletter
\def\Pa{\mathbf{Pa}}
\def\X{\mathbf{X}}
\def\Y{\mathbf{Y}}
\def\T{\mathrm{T}}
\def\mh{m^{h}}
\def\mc{m_{c}}
\def\mch{m_{c}^{h}}
\def\am{\alpha_{\mu}}
\def\aw{\alpha_{w}}
\def\bmu{\bolds{\mu}}
\def\bnu{\bolds{\nu}}
\newcommand{\eref}[1]{(\ref{#1})}
\makeatother

\begin{document}

\begin{frontmatter}

\title{Addendum on the scoring of Gaussian directed acyclic graphical
models}
\runtitle{Addendum on scoring Gaussian DAG models}

\begin{aug}
\author[a]{\fnms{Jack} \snm{Kuipers}\corref{}\ead[label=e1]{jack.kuipers@ur.de}},
\author[b]{\fnms{Giusi} \snm{Moffa}\ead[label=e2]{giusi.moffa@ukr.de}}
\and
\author[c]{\fnms{David} \snm{Heckerman}\ead[label=e3]{heckerma@microsoft.com}}
\runauthor{J. Kuipers, G. Moffa and D. Heckerman}

\affiliation{Regensburg University, Regensburg University and Microsoft
Research}

\address[a]{J. Kuipers\\
Institut f\"ur Theoretische Physik\\
Universit\"at Regensburg\\
D-93040 Regensburg\\
Germany\\
\printead{e1}}

\address[b]{G. Moffa\\
Institut f\"ur funktionelle Genomik\\
Universit\"at Regensburg\\
Josef Engertstra\ss e 9\\
93053 Regensburg\\
Germany \\
\printead{e2}}

\address[c]{D. Heckerman\\
Microsoft Research\\
1100 Glendon Ave Suite PH1\\
Los Angeles, California 90024\\
USA\\
\printead{e3}}
\end{aug}

\received{\smonth{2} \syear{2014}}

%
\begin{abstract}
We provide a correction to the expression for scoring Gaussian directed
acyclic graphical models derived in Geiger and Heckerman
[\textit{Ann. Statist.} \textbf{30} (2002) 1414--1440]
and discuss how to evaluate
the score efficiently.
\end{abstract}

%

\end{frontmatter}

Gaussian directed acyclic graph (DAG) models represent a particular
type of Bayesian networks where the node variables are assumed to come
from a multivariate Gaussian distribution. The Bayesian Gaussian
equivalent (BGe) score was introduced in \citeauthor{gh94}
(\citeyear{gh94,gh02}),
\citet{hg95} for
learning such networks.

For brevity, we omit formal definitions and refer the reader to \citet
{gh02}, while following their notation in considering DAG models $m$
with~$n$ nodes corresponding to the set of variables $\X= \{X_1,\ldots,X_n\}$. Let $\mh$ be the model hypothesis that the true distribution
of $\X$ is faithful to the DAG model $m$, meaning that it satisfies
only and all the conditional independencies encoded by the DAG. For a
complete random data sample $d=\{\mathbf{x}_{1},\ldots,\mathbf{x}_{N}\}$ with $N$
observations and a complete DAG model $\mc$, the marginal likelihood is
[\citeauthor{gh02} (\citeyear{gh02}), Theorem~2]
%
\begin{equation}
\label{bgescore} p\bigl(d\mid\mh\bigr) = \prod_{i=1}^{n}
\frac{p(d^{\Pa_i\cup\{X_i\}}\mid\mch
)}{p(d^{\Pa_i}\mid\mch)},
\end{equation}
where $\Pa_i$ are the parent variables of the vertex $i$ and $d^{\Y}$
is the data restricted to the coordinates in $\Y\subseteq\X$. The BGe
score is the posterior probability of $\mh$ which is proportional to
the marginal likelihood in \eref{bgescore} and the graphical prior; see
equation~(2) of \citet{gh02}.

Different DAGs which encode the same set of conditional independencies
are said to belong to an equivalence class. Along with ensuring that
all DAGs in the same equivalence class are scored equally, the
modularity of the score allows the steps in structure MCMC [\citet
{art:MadiganY95}] to be evaluated much more efficiently. Order MCMC
[\citet{art:FriedmanK2003}, on the related space of triangular matrices]
as well as the edge reversal move of \citet{art:GrzegorczykH2008} would
not be possible without it.

For Gaussian DAG models, the likelihood is a multivariate normal
distribution with mean $\bmu$ and precision matrix $W$. The need for
global parameter independence, so that the expression of the score in
\eref{bgescore} holds, implies that the prior distribution of $(\bmu,W)$ must be normal-Wishart [\citet{gh02}]. The parameter $\bmu$ is taken
to be normally distributed with mean $\bnu$ and precision matrix $\am
W$, for $\am>0$. $W$ is Wishart distributed with positive definite
parametric matrix~$T$ (the inverse of the scale matrix) and degrees of
freedom $\aw$, with $\aw>n-1$. As detailed in the supplementary material
[\citet
{kmh14}], one finds
%
\begin{eqnarray}
\label{pdYresult} &&p\bigl(d^{\Y}\mid\mch\bigr)
\nonumber
\\[-8pt]
\\[-8pt]
\nonumber
&&\qquad= \biggl(\frac
{\am}{N+\am}
\biggr)^{{l}/{2}} \frac{\Gamma_{l} ({(N+\aw-n+l)}/{2}
)}{\pi^{
{lN}/{2}}\Gamma_{l} ({(\aw-n+l)}/{2} )}\frac{\vert T_{\Y\Y}
\vert^{{(\aw-n+l)}/{2}}}{\vert R_{\Y\Y} \vert^{{(N+\aw-n+l)}/{2}}},
\end{eqnarray}
where $l$ is the size of $\Y$, $A_{\Y\Y}$ means selecting the rows and
columns corresponding to $\Y$ of a matrix $A$,
%
\begin{equation}
\label{Gammandef} \Gamma_{l} \biggl(\frac{x}{2} \biggr) =
\pi^{{l(l-1)}/{4}}\prod_{j=1}^{l}\Gamma
\biggl(\frac{x+1-j}{2} \biggr)
\end{equation}
is the multivariate Gamma function and
%
\begin{equation}
\label{matrixR} R = T + S_N + \frac{N\am}{(N+\am)} (\bnu- \bar
{\mathbf{x}} ) (
\bnu- \bar{\mathbf{x}} )^{\T}
\end{equation}
is the posterior parametric matrix involving
%
\begin{equation}
\label{Sdef} \bar{\mathbf{x}} = \frac{1}{N} \sum_{i=1}^{N}
\mathbf{x}_{i},\qquad S_N = \sum_{i=1}^{N}
(\mathbf{x}_i-\bar{\mathbf{x}} )
(\mathbf{x}_i-\bar{\mathbf{x}} )^{\T}
\end{equation}
the sample mean and sample variance multiplied by $(N-1)$.

The result in \eref{pdYresult} is identical to equation (18) of \citet
{gh02}, once some factors are cancelled, apart from the manner in which
the matrix elements are chosen. The result in \citet{gh02} replaces the
$T_{\Y\Y}$ and $R_{\Y\Y}$ by $T_{\Y}$ and $R_{\Y}$, where $A_{\Y
}=((A^{-1})_{\Y\Y})^{-1}$. Inverting the matrices before the elements
are selected and then inverting again [as in \citet{gh02}] we found
inconsistent behavior on simulated data.

We may further compare to equation (24) of \citet{hg95}, which with the
current notation becomes
%
\begin{equation}
\label{pdYresulthg95} p\bigl(d^{\Y}\mid\mch\bigr) = \biggl
(\frac{\am}{N+\am}
\biggr)^{{l}/{2}} \frac{\Gamma_{l} ({(N+\aw)}/{2} )}{\pi
^{{lN}/{2}}\Gamma
_{l} ({\aw}/{2} )}\frac{\vert T_{\Y\Y} \vert^{{\aw
}/{2}}}{\vert R_{\Y\Y} \vert^{{(N+\aw)}/{2}}}
\end{equation}
while incorrectly defining the $S_N$ in the $R$ in \eref{matrixR} as
the sample variance. However, the same terminology, with the correct
formula for $S_N$, is used in \citet{gh94} whose equation (12) is
otherwise identical to \eref{pdYresulthg95}.

The difference in the powers of the determinants between \eref
{pdYresult} and \eref{pdYresulthg95} could lead to a subtle, and hard
to predict, change in the scores. There is also the same loss of
$l$-dependence in the arguments of the multivariate gamma functions.
The ratio of gamma functions for each node now actually decreases with
$l$ while the ratio from~\eref{pdYresult} increases instead. As
discussed in the supplementary material [\citet{kmh14}], using \eref
{pdYresulthg95} instead of \eref{pdYresult} effectively penalises each
node with $l$ parents by a factor $\sim N^l$, giving a substantial bias
toward sparse DAGs. This bias is likely to be present in early works
implementing the score of \citet{hg95} and possibly remains in legacy code.
%

\begin{supplement}[id=suppA]
\stitle{Deriving and simplifying the BGe score}
\slink[doi]{10.1214/14-AOS1217SUPP} 
\sdatatype{.pdf}
\sfilename{aos1217\_supp.pdf}
\sdescription{We detail the steps used to derive \eref{pdYresult} and
simplify the ratios appearing in \eref{bgescore} to improve the
numerical computation of the score.}
\end{supplement}

%

%


\end{document}